%% file: main.tex
\PassOptionsToPackage{table}{xcolor}
\documentclass[letterpaper]{article}
\usepackage[preprint]{aaai2027}

\usepackage[hyphens]{url}
\usepackage{graphicx}
\usepackage{natbib}
\usepackage{caption}
\usepackage{booktabs}
\usepackage{amsmath}
\usepackage{amssymb}
\usepackage{fontawesome5}
\usepackage{mathtools}
\usepackage{xfrac}
\usepackage{multirow}
\usepackage{bm}
\usepackage{pifont}
\usepackage[capitalize,noabbrev]{cleveref}
\usepackage{tikz}
\usetikzlibrary{arrows.meta}
\PassOptionsToPackage{table}{xcolor}
\newcommand{\singlerightarrow}{%
  \tikz[baseline=-0.6ex]
    \draw[-{Stealth[length=3pt,width=3pt]},
          very thick]
      (0,0) -- (1.5em,0);
}

\newcommand{\doublerightarrow}{%
  \tikz[baseline=-0.6ex]
    \draw[-{Stealth[length=3pt,width=3pt]
            . Stealth[length=3pt,width=3pt]},
          very thick]
      (0,0) -- (1.5em,0);
}

\newcommand{\triplerightarrow}{%
  \tikz[baseline=-0.6ex]
    \draw[-{Stealth[length=3pt,width=3pt]
            . Stealth[length=3pt,width=3pt]
            . Stealth[length=3pt,width=3pt]},
          very thick]
      (0,0) -- (1.5em,0);
}

\makeatletter
\let\old@fnsymbol\@fnsymbol
\renewcommand{\@fnsymbol}[1]{%
  \ifcase#1%
  \or \faHandshake
  \or \faEnvelope
  \else \old@fnsymbol{#1}%
  \fi
}
\makeatother

\title{MLAfication: Efficiently Enabling High-Fidelity MLA for Modern LLMs}
\author{
Zile Wang\equalcontrib,
\quad Zhide Xie\equalcontrib,
\quad Tao Ji\corresponding
}
\affiliations{
Fudan University\\
\texttt{\{zilewang24@m., xiezd24@m., taoji@\}fudan.edu.cn}
}

\begin{document}
\maketitle
\makeatletter
\let\@fnsymbol\AAAIorigfnsymbol
\makeatother
\begin{abstract}
    \input{sec/1_abstract}

\end{abstract}

\input{sec/2_intro}
\input{sec/3_method}
\input{sec/4_experiment}
\input{sec/5_related_work}
\input{sec/6_conclusion}

\bibliography{main}

\clearpage
\appendix
\twocolumn[
\begin{center}
    {\LARGE\bfseries Supplementary Material\par}
    \vspace{1.5em}
\end{center}
]

\input{sec/8_appendix}

\end{document}

%% file: sec/1_abstract.tex

Multi-Head Latent Attention (MLA) compresses the KV cache
into a latent representation, reducing inference memory and bandwidth.
Existing MHA2MLA/TransMLA methods do not support QKNorm and require
costly recovery on modern LLMs, especially MoE. We present
\textbf{MLAfication}, a QKNorm-compatible, parameter-efficient extension
of MHA2MLA. It combines calibrated KNorm, parameter-efficient
distillation, contribution-aware Partial-RoPE, and activation-aware
joint KV initialization. 
Across MHA and GQA models, including QKNorm- and MoE-equipped variants,
MLAfication preserves performance under substantial KV-cache compression,
surpasses the current SOTA on commonsense and long-context
benchmarks, and extends MLA conversion to modern architectures. 
On Qwen3-30B-A3B, MLAfication delivers up to $5.35\times$ the serving
throughput of BF16 GQA and, combined with 4-bit KV quantization, reduces
the KV-cache footprint by $94.73\%$.
Our code is available at \url{https://github.com/JT-Ushio/MLAfication}.


%% file: sec/2_intro.tex
\section{Introduction}
\label{sec:intro}



The rapid scaling of LLMs has led to increasing challenges in inference efficiency, particularly due to the substantial memory and bandwidth costs of Key-Value (KV) caches. 
To address this bottleneck, DeepSeek introduced Multi-Head Latent Attention (MLA, \citeyear{corr/abs-2405-04434}), which compresses KV representations into a compact latent space while preserving expressive attention capabilities. 
MLA has since been adopted by several large-scale foundation models, including DeepSeek-V2/V3/R1/V3.2~\cite{corr/abs-2512-02556}, Kimi K2/K3~\cite{kimiteam2026kimik3openfrontier}, and the GLM-5 series~\cite{glm5team2026glm5vibecodingagentic}, highlighting its potential as an efficient attention architecture for modern LLMs.

MHA2MLA~\cite{ji-etal-2025-towards} first demonstrated that LLMs originally optimized for MHA (\citeyear{nips/VaswaniSPUJGKP17}) or GQA (\citeyear{emnlp/AinslieLJZLS23}) can be efficiently adapted to MLA inference without retraining from scratch. 
By introducing \ding{182}\textit{Partial-RoPE} and \ding{183}\textit{KV-joint low-rank approximation}, MHA2MLA enables data-efficient conversion and effectively recovers performance.
TransMLA~\cite{conf/nips/MengTYSZ25} further improved Partial-RoPE variants (RoRoPE and FreqFold) and KV balancing, delivering stronger zero-shot performance and more efficient fine-tuning.
Beyond text-only LLMs, the paradigm has recently been extended to other model families, including vision-LMs~\cite{conf/aaai/FanSJSG26}, ASR models~\cite{zhang2026whisper}, and autoregressive video diffusion models~\cite{yesiltepe2026videomlalowranklatentkv}, demonstrating its broader applicability.

However, extending MHA2MLA to modern LLMs remains challenging due to incompatibilities with widely adopted architectural components, particularly QKNorm (\citeyear{henry2020query}) and Mixture-of-Experts (MoE,~\citeyear{shazeer2017outrageously}).
While QKNorm stabilizes attention computation by mitigating logit explosion, its normalization conflicts with MLA's latent projection mechanism~\cite{team2025kimi,anson2025controlling}.
Similarly, MoE improves model scalability but makes full-parameter MHA2MLA vulnerable to disrupting the expert load balancing learned during pre-training.

We introduce \textbf{MLAfication}, extending
MHA2MLA~\citep{ji-etal-2025-towards} to QKNorm and MoE architectures.
It resolves QKNorm's incompatibility with efficient MLA decoding through
calibrated KNorm and recovers performance through parameter-efficient
distillation. We further refine MHA2MLA's Partial-RoPE and low-rank
approximation stages using MHA2MLA-VLM~\citep{conf/aaai/FanSJSG26} and
SVD-LLM V2~\citep{wang-etal-2025-svd-llm}, respectively.

\paragraph{Contributions.}
\begin{itemize}
    \item We enable standard MLA decoding for QKNorm models through
    calibrated KNorm, without extra cache or custom kernels.

    \item QKV-only distillation trains $1.6\%$--$11.1\%$ of model
    parameters while matching full fine-tuning and limiting MoE routing drift.

    \item Across four model families, MLAfication improves both commonsense and long-context accuracy over SOTA and delivers up to
    $5.35\times$ serving throughput.
\end{itemize}


%% file: sec/3_method.tex
\section{MLAfication}
\label{sec:mlafication}

\begin{figure*}[t]
    \centering
    \includegraphics[width=0.9\textwidth]{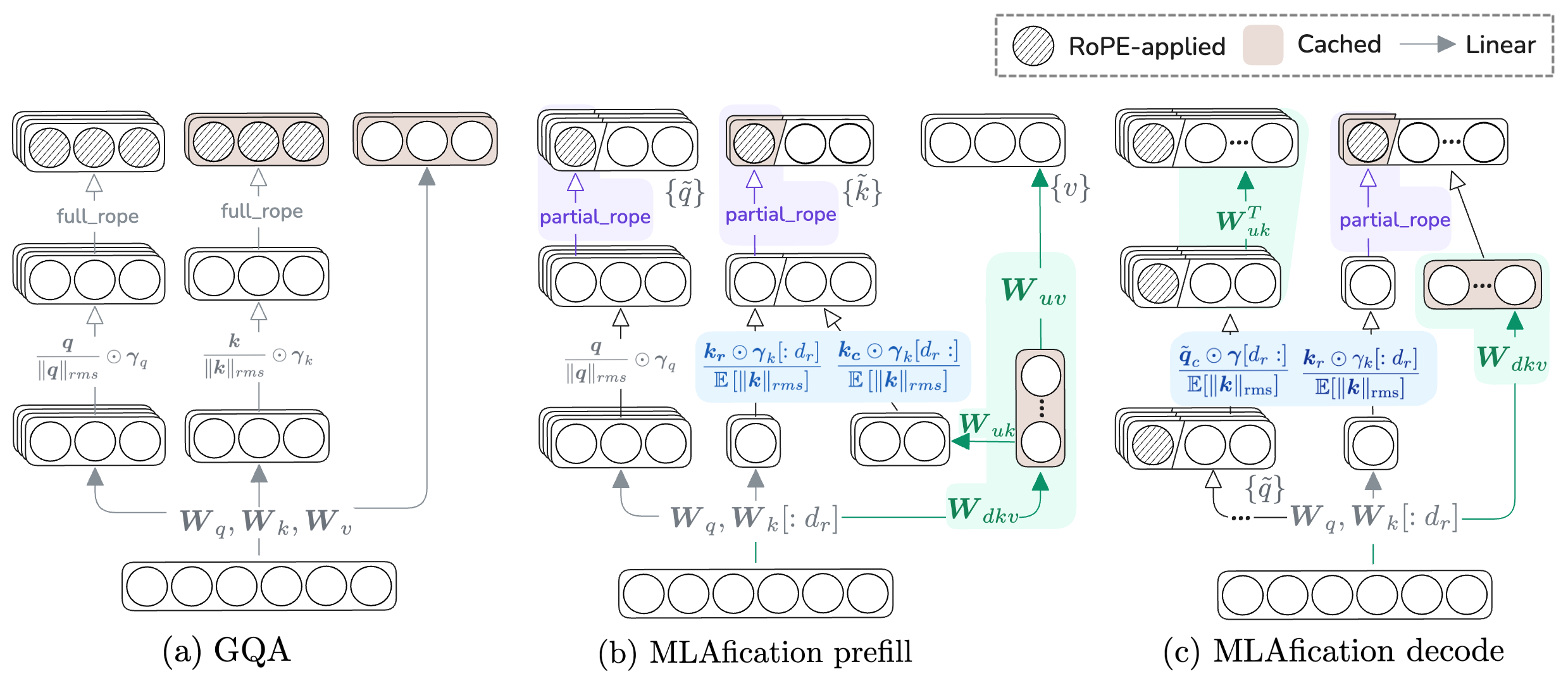}
    \vspace{-0.2cm}
    \caption{
        Adapting QKNorm to efficient MLA decoding. For compactness in the schematic, we write $\bm q_c$ and
$\bm q_r$ for $\bm q_{i,\mathrm{nope}}$ and
$\bm q_{i,\mathrm{rope}}$, respectively, and use the same convention
for $\bm k$.
    }
    \label{fig:qknorm}
\end{figure*}

MLAfication builds on MHA2MLA (\citeyear{ji-etal-2025-towards}) and extends it in three key respects\footnote{Given space constraints and the inherent complexity of both the MLA and the MHA2MLA paradigm, we defer their detailed background to
the Appendix C. Readers unfamiliar with either are encouraged to
consult the original papers or the Appendix C.}.
\S\ref{ssec:qknorm_mla} resolves the incompatibility between QKNorm and efficient MLA decoding, enabling the MLAfication of MHA/GQA models equipped with QKNorm (\Cref{fig:qknorm}).
\S\ref{ssec:peft_and_distill} introduces parameter-efficient distillation for low-cost architecture editing.
\S\ref{ssec:mla_initialization} improves the two initialization steps by adopting the Partial-RoPE strategy from MHA2MLA-VLM (\citeyear{conf/aaai/FanSJSG26}) and the activation-aware low-rank approximation from SVDLLM-V2 (\citeyear{wang-etal-2025-svd-llm}).

\subsection{QKNorm Compatibility}
\label{ssec:qknorm_mla}

\paragraph{Why Does QKNorm Prevent MLA Decoding?}

Modern LLMs such as Qwen3~\citep{yang2025qwen3} apply
RMSNorm~\citep{zhang2019root} to each query and key as
$\hat{\bm x}_i^{(h)}
=\bm\gamma_x^{(h)}\odot\bm x_i^{(h)}
 /\left\|\bm x_i^{(h)}\right\|_{\operatorname{rms}}$, where $\bm{x}\in\{\bm{q},\bm{k}\}$,
$\bm\gamma_x^{(h)}$ is a learnable element-wise gain, and
$\left\|\cdot\right\|_{\operatorname{rms}}$ denotes the root-mean-square magnitude. QKNorm stabilizes training by
controlling attention-logit magnitudes. Its key-side normalization,
however, conflicts with the matrix absorption used in standard
efficient MLA decoding~\citep{anson2025controlling,team2025kimi}.

After Partial-RoPE conversion and joint KV compression, the pre-RoPE
key of head $h$ is reconstructed as
\begin{equation}
    \bm k_j^{(h)}
    =
    \begin{bmatrix}
        \bm W_{uk}^{(h)}\bm c_{j,kv} \\
        \bm W_{kr}^{(h)}\bm x_j
    \end{bmatrix},
    \qquad
    \bm c_{j,kv}
    =
    \bm W_{dkv}\bm x_j,
\label{eq:mla_key}
\end{equation}
where $\bm c_{j,kv}$ is the compressed KV representation. Without
KNorm, the NoPE attention logit can be rewritten as
\begin{equation}
    \hat{\bm q}_{i,\mathrm{nope}}^{(h)\top}
    \bm W_{uk}^{(h)}
    \bm c_{j,kv}
    =
    \left(
        \bm W_{uk}^{(h)\top}
        \hat{\bm q}_{i,\mathrm{nope}}^{(h)}
    \right)^\top
    \bm c_{j,kv}.
\label{eq:mla_absorption}
\end{equation}
Thus, $\bm W_{uk}^{(h)}$ can be absorbed into the query projection, and
attention can operate directly on the latent KV cache.

KNorm breaks this absorption because its normalization factor depends
on both components of the reconstructed key:
\begin{equation}
    \left\|\bm k_j^{(h)}\right\|_{\operatorname{rms}}
    =
    \left\|
        \begin{bmatrix}
            \bm W_{uk}^{(h)}\bm c_{j,kv} \\
            \bm W_{kr}^{(h)}\bm x_j
        \end{bmatrix}
    \right\|_{\operatorname{rms}}.
\label{eq:knorm_denominator}
\end{equation}
While the learned scale $\bm\gamma_k^{(h)}$ is static and can be folded
into a projection, $\left\|\bm k_j^{(h)}\right\|_{\operatorname{rms}}$ varies across
tokens. It therefore cannot be absorbed into a fixed
query-side projection. Preserving exact KNorm would require either
materializing the reconstructed keys or caching a normalization factor
for every token and KV head and applying it with a specialized
attention kernel. Either option departs from standard latent-cache MLA
decoding.

QNorm remains compatible with MLA because only the current query is
materialized during autoregressive decoding. It can therefore be
normalized. We retain QNorm
unchanged and adapt only KNorm.

\paragraph{Calibrated KNorm approximation.}

Removing KNorm would alter the key scale inherited from pretraining.
We instead replace the token-dependent
$\left\|\bm k_j^{(h)}\right\|_{\operatorname{rms}}$ with its calibration average, using
one constant per layer and KV head. Omitting the layer index for
clarity, we then define
$
    \alpha_h
    =
    \mathbb{E}
    \left[
        \left\|\bm k_j^{(h)}\right\|_{\operatorname{rms}}
    \right],
$
where $\boldsymbol{k}_j^{(h)}$ belongs to a calibration sample set. 
We then approximate the normalized key as
$\hat{\bm k}_j^{(h)}
\approx
\bm\gamma_k^{(h)}\odot\bm k_j^{(h)}/\alpha_h$.

Since both $\alpha_h$ and $\bm\gamma_k^{(h)}$ are token-independent,
their NoPE component can be folded into the key up-projection:
\begin{equation}
    \bm W_{uk}^{\prime(h)}
    =
    \operatorname{Diag}\!\left(
            \bm\gamma_{k,\mathrm{nope}}^{(h)}
    \right)
    \cdot\frac{1}{\alpha_h}\cdot\bm W_{uk}^{(h)} .
\end{equation}
The NoPE logit can therefore be computed directly from the latent
cache:
\begin{equation}
    \hat{\bm q}_{i,\mathrm{nope}}^{(h)\top}
    \bm W_{uk}^{\prime(h)}
    \bm c_{j,kv}
    =
    \left(
        \bm W_{uk}^{\prime(h)\top}
        \hat{\bm q}_{i,\mathrm{nope}}^{(h)}
    \right)^\top
    \bm c_{j,kv}.
\end{equation}
The same static scaling is folded into the RoPE key projection before
rotation. This restores standard MLA decoding without extra cache
states or custom normalization kernels. During recovery training, the
edited projections and QKNorm parameters are jointly refined.

\begin{table}[h]

\centering
\small
\begin{tabular}{lccccc}
\toprule
 & 1K & 2K & 4K & 8K & 16K \\
\midrule
$\alpha_h$ & 0.8243 & 0.8185 & 0.8228 & 0.8260 & 0.8305 \\
\bottomrule
\end{tabular}
\caption{Stability of $\alpha_h$ across context lengths, estimated on a calibration set of 64 samples.}
\label{tab:qknorm_ratio}
\end{table}
In a preliminary experiment, we examine whether $\alpha_h$ is sensitive to context length. We select the middle KV
head in the middle layer\footnote{The same stability is observed across all layers and heads. } of Qwen3-8B and estimate $\alpha_h$ using
sequences ranging from 1K to 16K. 
Table~\ref{tab:qknorm_ratio} shows that the estimates remain within a narrow
range of 0.8185\textasciitilde0.8305, with a coefficient of variation of only $0.48\%$.
This stability supports using a single calibrated constant in place of the token-wise KNorm denominator.

\begin{figure}[t]
    \centering
    \includegraphics[width=\columnwidth]{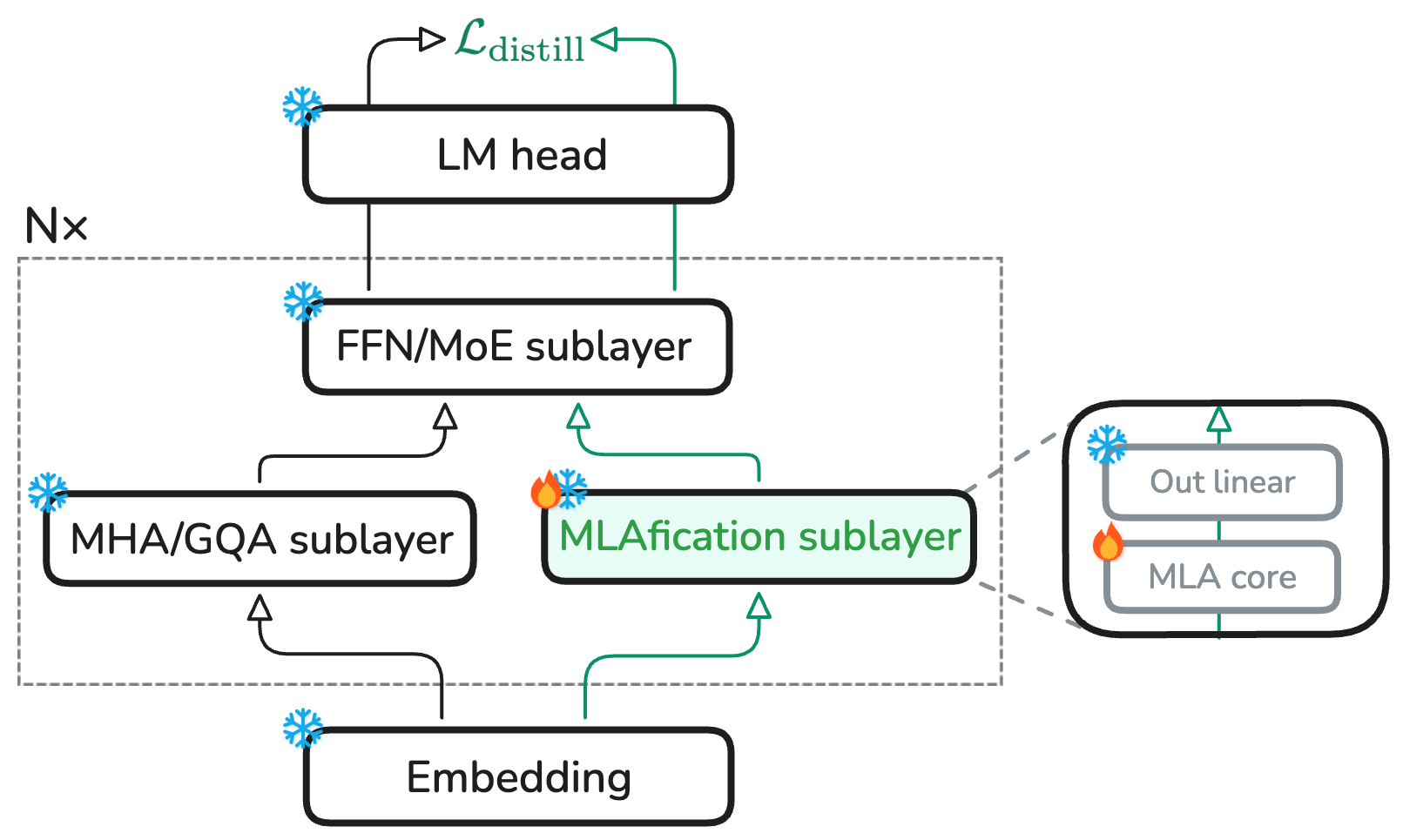}
    \vspace{-0.5cm}
    \caption{
        Parameter-Efficient Distillation.
        Snowflakes and flames denote frozen and trainable modules,
        respectively.
    }
    \label{fig:peft}
\end{figure}
\subsection{Parameter-Efficient Distillation}
\label{ssec:peft_and_distill}
Full-parameter recovery incurs substantial compute and memory overhead and may unnecessarily perturb pretrained representations, including MoE routing and expert specialization. As shown in \Cref{fig:peft}, we instead combine \textit{distillation}
with \textit{parameter-efficient} fine-tuning (PEFT), updating only the parameters
modified by MLAfication.

\paragraph{Distillation.}

The original MHA/GQA model serves as the teacher and the MLAfied model as the student.
Given the same sequence, we minimize the KL divergence from the teacher distribution to the student distribution:
\begin{equation}
\mathcal L_{\mathrm{distill}}
=
\frac{1}{L}
\sum_{t=1}^{L}
p_T(x_t\mid x_{<t})
\log
\frac{
p_T(x_t\mid x_{<t})
}{
p_S(x_t\mid x_{<t})
},
\label{eq:distillation}
\end{equation}
where $p_T$ and $p_S$ denote the teacher and student output distributions.
Unlike one-hot next-token targets, the teacher distribution provides dense supervision and conveys implicit knowledge.
It improves data efficiency during MLAfication.

\paragraph{PEFT.}
We update only the Q/K/V-side parameters modified by MLAfication, including the latent down- and up-projections, the projections associated with the RoPE-preserved subspace, and the corresponding QKNorm parameters. The attention output projections, embeddings, LM head, and
all FFN parameters remain frozen.

For modern MoE models, we also freeze all routers and experts. This preserves
the load-balancing behavior learned during pretraining. Distillation
further limits routing changes caused indirectly by altered attention
representations.
Freezing the backbone removes its gradient and optimizer-state costs
and allows the teacher and student to share the same frozen parameter
storage. This leaves only $0.91$B of $8.19$B parameters ($-88.9\%$) trainable in Qwen3-8B and $0.50$B of $30.53$B parameters ($-98.4\%$) trainable in Qwen3-30B-A3B.

\subsection{Refining MHA2MLA's \ding{182}\&\ding{183} Approximation}
\label{ssec:mla_initialization}

\paragraph{\ding{182} Partial-RoPE.}

MLA retains a small RoPE component for positional encoding while compressing
the remaining NoPE key dimensions. We adopt the contribution-aware
Partial-RoPE strategy from MHA2MLA-VLM~\citep{conf/aaai/FanSJSG26}, which
retains the rotary subspaces that most strongly influence the attention
distribution. Since we adopt this procedure without modification, we omit its details here
for brevity and refer readers to the original paper.

\paragraph{\ding{183} Activation-aware low-rank approximation.}

Let $\bm W_{\bar{k}}$ denote the NoPE key
projection. We stack it with the value projection as
$\bm W_{kv}=[\bm W_{\bar{k}};\bm W_v]$. Vanilla truncated
SVD minimizes the weight-space error
$\min_{\operatorname{rank}(\bm W_{kv}')\le d_{kv}}
\|\bm W_{kv}-\bm W_{kv}'\|_F^2$, ignoring the input distribution.

We instead adopt the loss-optimized objective of SVDLLM-V2~\citep{wang-etal-2025-svd-llm}. Given calibration activations
$\bm X\in\mathbb R^{d\times N}$, it minimizes
\begin{equation}
    \min_{\operatorname{rank}(\bm W_{kv}')\le d_{kv}}
    \left\|
        \bm W_{kv}\bm X-\bm W_{kv}'\bm X
    \right\|_F^2.
\label{eq:activation_reconstruction}
\end{equation}

We first eigendecompose the activation second-moment matrix as
$\bm X\bm X^\top=\bm U_x\bm\Lambda_x\bm U_x^\top$ and form
$\bm D=\bm W_{kv}\bm U_x\bm\Lambda_x^{1/2}$. Let
$\bm D\approx\bm U_r\bm\Sigma_r\bm V_r^\top$ be its rank-$d_{kv}$
truncated SVD. We initialize the MLA factors as
\begin{equation}
\begin{aligned}
    \bm W_{dkv}
    =
    \bm\Sigma_r^{1/2}
    \bm V_r^\top
    \bm\Lambda_x^{-1/2}
    \bm U_x^\top, \quad
    \begin{bmatrix}
        \bm W_{uk} \\
        \bm W_{uv}
    \end{bmatrix}
    =
    \bm U_r\bm\Sigma_r^{1/2},
\end{aligned}
\label{eq:svdllm_v2_initialization}
\end{equation}
where $\bm\Lambda_x^{-1/2}$ is evaluated over the nonzero eigenspace.
Their product gives the rank-$d_{kv}$ approximation that minimizes
\Cref{eq:activation_reconstruction}.
This initialization minimizes KV output reconstruction error on the
calibration rather than naive SVD.

%% file: sec/4_experiment.tex
\section{Experiments}
\label{sec:exper}

\input{tab/commonsense_qwq}

We evaluate four model families\footnote{All models are \texttt{Base} models instead of \texttt{Instruct} ones.}: MHA (SmolLM\textsubscript{\textit{1.7B}}), GQA (Qwen2.5\textsubscript{\textit{7B}}), GQA+QKNorm (Qwen3\textsubscript{\textit{8B}}), and GQA+QKNorm+MoE (Qwen3\textsubscript{\textit{30B-A3B}}).
Following prior work~\citep{ji-etal-2025-towards,conf/nips/MengTYSZ25},
we source the fine-tuning data from SmolLM
training corpus~\citep{allal2024smollm}.
We evaluate model performance on Commonsense, which comprises six sub-datasets, and LongBench v1~\citep{bai2024longbench}. We adopt continued pre-training (CPT) as a baseline to eliminate performance differences caused by the training data, since our fine-tuning corpus differs from the proprietary data used to train Qwen. We also compare against TransMLA, the current SOTA method. Note that TransMLA does not apply to models equipped with QKNorm.
Unless otherwise specified, MLAfication use contribution-aware partial-RoPE, SVDLLM V2, calibrated KNorm approximation when QKNorm is present, and parameter-efficient distillation.
Full hyperparameters are provided in Appendix A.

\subsection{Main Results}

Table~\ref{tab:main} reports the main evaluation across four model families.
Importantly, the data-control experiment reveals the SmolLM-derived dataset is
effective: CPT outperforms the original models in 17/21 settings for the Qwen series. Comparing MLAfication against CPT is therefore
essential, as comparisons only against the original models would incorrectly attribute data-induced improvements to our method.


First, MLAfication and the current SOTA TransMLA can be fairly compared across both MHA and GQA families. On commonsense (CS) benchmarks, MLAfication underperforms TransMLA on MHA‑based models by an average of 0.21 points, but outperforms it on GQA‑based models by an average of 0.68 points. On LongBench (LB), MLAfication achieves better results in 5/6 evaluation settings, highlighting its effectiveness in long‑context scenarios. Given that GQA—as a refinement of MHA—is increasingly preferred in architectural choices, and that long‑context capabilities are becoming more critical in multimodal and agent‑based applications, our results suggest that MLAfication offers greater practical value than the current SOTA TransMLA.

Second, the results hold on recent mainstream architectures that use QKNorm and MoE.
With calibrated KNorm approximation, MLAfication resolves the structural conflict between QKNorm and MLA-style matrix absorption, enabling conversion of Qwen3-8B and Qwen3-30B-A3B.
On Qwen3-8B, even the strongest compression ratio keeps the CS and LB drops from CPT at 3.35 and 2.12 points, respectively. This degradation is comparable to Qwen2.5-7B, showing that our QKNorm adaptation preserves the recoverability of MLAfication on normalized attention models.
On Qwen3-30B-A3B, at $-68.75\%$ KV reduction, CS and LB are 1.60 and 0.13 points below CPT, respectively. At the strongest compression ratio, the CS and LB gaps remain 4.05 and 2.04 points, respectively.
Together, these results show that MLAfication successfully adapts to the QKNorm-based GQA and MoE architectures that are dominant in recent LLMs.

Third, consistent with prior findings, more aggressive KV-cache
compression leads to greater performance degradation because it
introduces larger approximation errors. Moreover, GQA already reduces
the KV cache through KV-head sharing, leaving less room for further
compression than MHA. 

Overall, the main results show that MLAfication can be applied to recent mainstream LLM architectures including GQA, QKNorm, and MoE, while minimizing quality degradation under KV-cache compression.

\begin{figure}[!t]
\centering
\includegraphics[width=\linewidth]{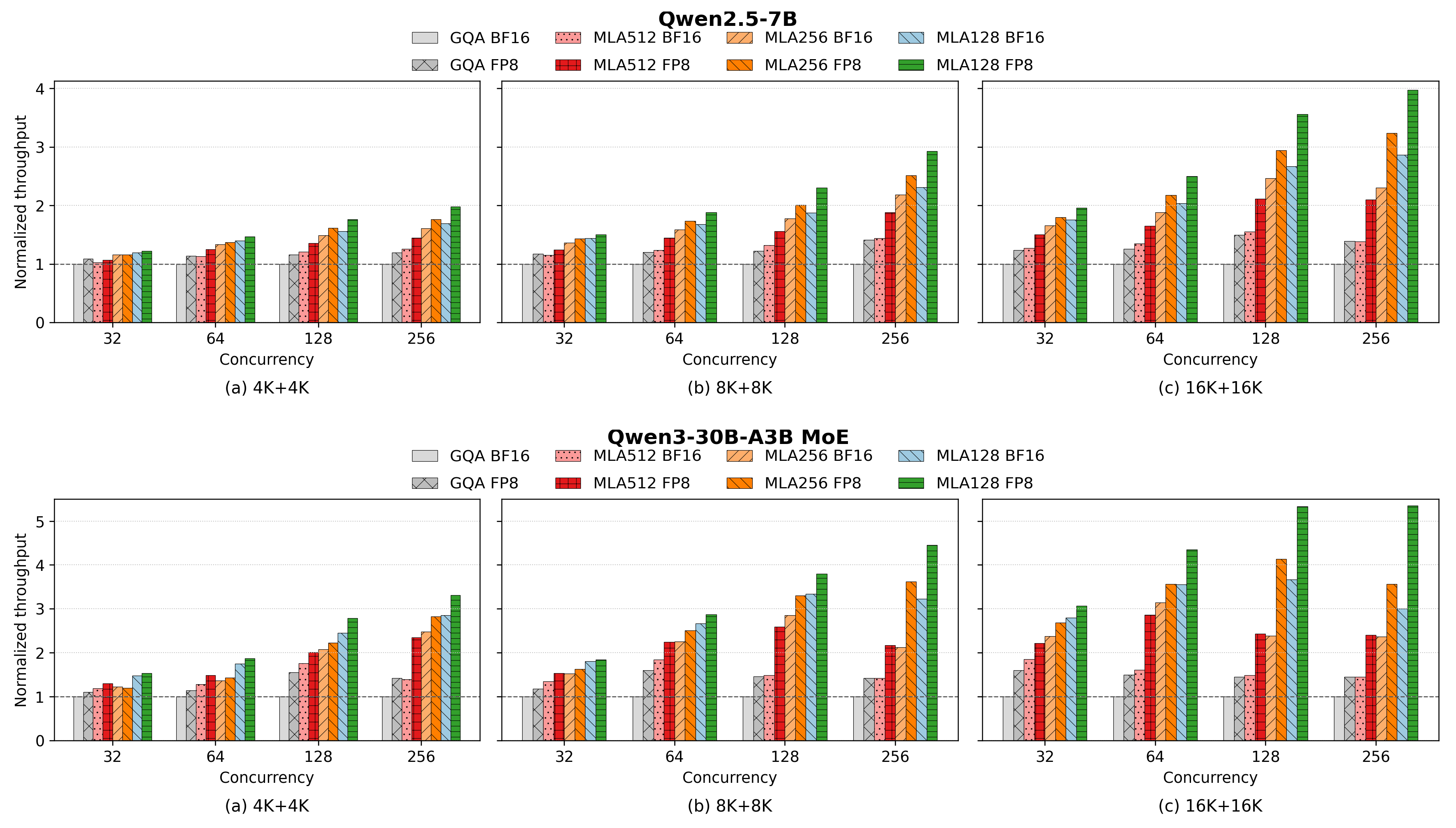}
\vspace{-0.5cm}
\caption{
Normalized inference throughput across prompt lengths and concurrency levels.
Throughput is normalized to the baseline under the same setting; the dashed horizontal line denotes the baseline.
}
\label{fig:throughput}
\end{figure}

\subsection{Serving Efficiency}

We evaluate end-to-end serving efficiency with vLLM on a single NVIDIA
H200 GPU, using \texttt{BF16} model weights with either \texttt{BF16} or \texttt{FP8\_E4M3} KV cache.
Following the evaluation protocol of TransMLA,
we use equal input and output lengths: 4K+4K, 8K+8K, and 16K+16K. We measure output-token throughput under concurrency levels of 32, 64, 128, and 256. Figure~\ref{fig:throughput} reports throughput normalized by the corresponding \texttt{BF16} GQA baseline under the same workload.
Evaluation settings are provided in Appendix B.

We adapt optimized inference kernels to multiple latent
dimensions, $d_c \in \{128, 256, 512\}$, while fixing the RoPE dimension
to $d_r=64$, and support both \texttt{BF16} and \texttt{FP8\_E4M3} formats.
To our knowledge, TransMLA evaluates the performance of the service only for a $d_c=512$ workload setting with $d_r=64$, and does not report \texttt{the FP8} KV-cache.

Figure~\ref{fig:throughput} shows the largest gains generally occur
in more memory-intensive serving regimes, especially for lower-dimensional
MLA with \texttt{FP8} KV caches.
On Qwen2.5-7B at 16K+16K and concurrency 256, \texttt{BF16} MLA-512/256/128 achieves $1.38\times$/$2.30\times$/$2.87\times$ normalized throughput, while \texttt{FP8} MLA-128 reaches $3.98\times$.The gains also transfer to the MoE architecture,
under the same workload, \texttt{BF16} MLA-512/256/128 reaches $1.45\times$/$2.37\times$/$3.00\times$ on Qwen3-30B-A3B, and \texttt{FP8} MLA-128 further increases throughput to $5.35\times$ over the \texttt{BF16} GQA baseline.

It shows that MLAfication not only improves serving efficiency
across both dense and MoE, but also composes effectively
with KV-cache quantization, with their combination yielding the strongest
throughput gains.

\input{tab/qk_norm}

\begin{figure}[t]
\centering
\includegraphics[width=\linewidth]{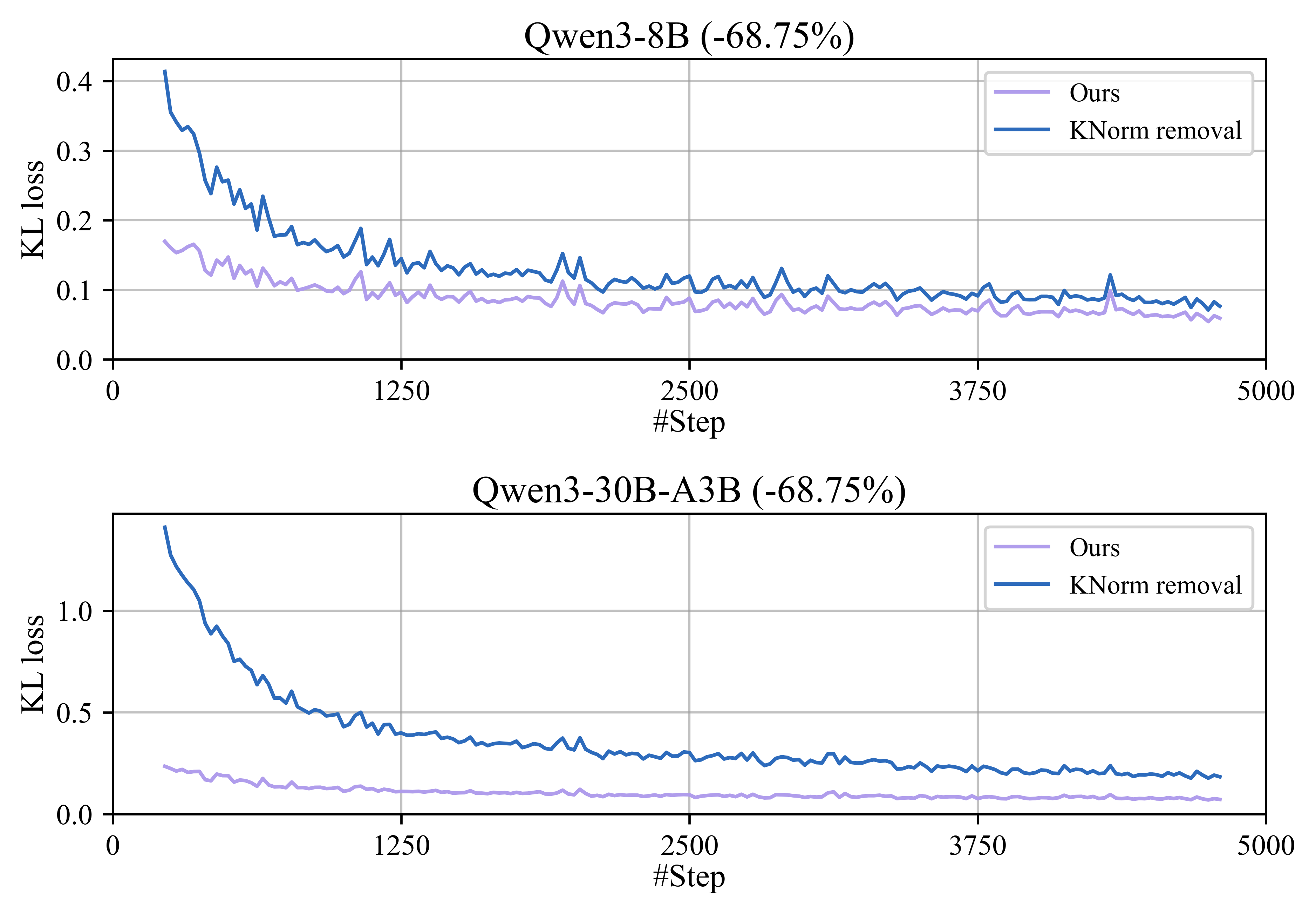}
\vspace{-0.8cm}
\caption{
KL-divergence loss curves for QKNorm ablation.
Our method achieves lower loss and faster convergence than directly removing KNorm.
}
\label{fig:qknorm_loss}
\end{figure}

\input{tab/quant_30B}

\subsection{QKNorm Compatibility}
\label{ssec:exp_qknorm}

QKNorm blocks standard MLA matrix absorption unless the KNorm denominator is handled explicitly.
We compare calibrated KNorm approximation with direct KNorm removal.
As shown in Table~\ref{tab:qknorm_ablation}, removing KNorm degrades downstream performance under all compression ratios.
On Qwen3-8B, our method improves CS by 1.3\% to 1.5\% and LB by 0.8\% to 3.6\% over KNorm removal. On Qwen3-30B-A3B, the improvement is more pronounced, with CS gains of 5.6\% to 11.8\% and LB gains of 13.5\% to 42.9\%.
The loss curves in Figure~\ref{fig:qknorm_loss} further show that our method converges faster and reaches lower distillation loss.
These results show that our method preserves conversion-induced degradation while keeping the attention computation compatible with latent caching.

\subsection{Integration with KV-Cache Quantization}

MLAfication and KV-cache quantization act on different components of cache size and can therefore be combined.
As shown in Table~\ref{tab:30b-a3b-quant-kv}, adding $r=128$ MLAfication to \texttt{Int4\_Quanto} increases the KV-cache reduction from 71.88\% to 94.73\% while still achieving 27.21 on LongBench.
At $r=256$, \texttt{Int4\_Quanto} scores 28.89 with a 91.21\% reduction, while quantization-only \texttt{Int2\_HQQ} scores 28.97 with an 84.38\% reduction.
The combined configuration therefore reduces the remaining cache by a further 43.75\% with a difference of only 0.08 LongBench points.

\subsection{Ablation Studies}

\subsubsection{Parameter-Efficient Distillation}

We ablate two key choices in recovery training: the training objective and the trainable parameter set (Table~\ref{tab:ablation_distill}).

\input{tab/ablation_at_distill}

First, logit distillation provides more effective recovery supervision than next-token prediction (NTP).
It improves CS in 9 of the 12 matched comparisons and LB in all 12, with larger and more consistent gains on LongBench.
At the strongest compression ratio under full-parameter tuning, distillation improves LB by 3.0 points on SmolLM-1.7B and 4.1 points on Qwen3-8B.
Unlike NTP's single hard target, distillation transfers the teacher's full output distribution and better constrains the deviation introduced by MLAfication.

Second, QKV-only fine-tuning matches full-parameter tuning closely under the same recovery budget. Averaged over all matched CS and LB comparisons, its score differs from full-parameter tuning by only $-0.015$ points, while requiring substantially fewer trainable parameters and optimizer states. We therefore adopt logit distillation with QKV-only fine-tuning as our default recovery configuration.

\subsubsection{Low-Rank KV Initialization}

Table~\ref{tab:step0_attn_output_mse} compares low-rank KV initialization methods before recovery training using layer-wise attention-output MSE on Qwen2.5-7B.

\input{tab/step0_attn_output_mse}

SVDLLM V2 consistently yields substantially lower attention-output MSE than vanilla SVD. For example, at 68.75\% compression, it reduces the MSE from 482.43 to 86.58, with the advantage remaining pronounced under stronger compression. This shows that activation-aware reconstruction better preserves the original attention behavior and provides a stronger initialization for recovery training.

\subsection{Diagnostics and Practical Considerations}

\paragraph{Training-free conversion is only a diagnostic.}

Table~\ref{tab:zero-shot} shows training-free performance provides limited evidence of practical effectiveness. Although both methods retain a substantial fraction of the base models' short-context accuracy, their LongBench scores drop to near zero on both LLaMA-2-7B and Qwen2.5-7B, precisely in the long-context regime where KV-cache compression is most relevant. One possible explanation is that modification of RoPE weakens positional discriminability, which is more critical for long-context tasks than for short-context benchmarks. We therefore report training-free results only as diagnostics and base our main comparison on performance after recovery stage.

\input{tab/zero_shot_commonsense}

\input{tab/moe_load_balance}

\paragraph{MoE routing.}
\label{ssec:moe_load_balance}
For MoE models, attention conversion could in principle shift hidden-state distributions and disturb expert routing.
As shown in Table~\ref{tab:moe_load_balance}, MLAfication does not exhibit an increasing trend in load-balancing violations during recovery stage: across all compression ratios, the mean violation rate remains stable and is only 4\%--5\% above the HF baseline.
In contrast, KNorm removal raises mean violations by 25\% -- 63\% above the baseline.
This provides additional evidence that our method avoids substantial routing drift and preserves the pretrained MoE behavior.

%% file: tab/commonsense_qwq.tex
\begin{table*}[ht!]

\centering
\small
\begin{tabular}{lrclccccccl}

\toprule

\textbf{Model} & \textbf{Tokens} & \textbf{KV Mem.} & \textbf{Avg@$\to$} & \textsubscript{\textbf{MMLU}} & \textsubscript{\textbf{ARC}} & \textsubscript{\textbf{PIQA}} & \textsubscript{\textbf{HS}} & \textsubscript{\textbf{OBQA}} & \textsubscript{\textbf{WG}} & \textbf{Avg@LB}\\

\midrule
\rowcolor{blue!10} \multicolumn{11}{c}{\textit{\textbf{MHA foundation models}}} \\
\rowcolor{gray!10} \textbf{SmolLM-1.7B} & \textcolor{gray}{1T} &  \textcolor{gray}{--} & \textcolor{gray}{55.90} & \textcolor{gray}{39.27} & \textcolor{gray}{59.87} & \textcolor{gray}{75.73} & \textcolor{gray}{62.93} & \textcolor{gray}{42.80} & \textcolor{gray}{54.85} & \textcolor{gray}{18.66}\\
\arrayrulecolor{gray!30}
\textit{+ CPT} & \multirow{3}{*}{300M} & -- & 55.42 & 38.68 & 58.79 & 76.12 & 61.33 & 42.80 & 54.78  & 16.76\\
\textit{+ TransMLA} &  & -68.75\%   & \textbf{55.24}$_{\text{-0.18}}$&38.60 & 58.95 & 74.97 & 61.52 & 43.00 & 54.38 & 15.33$_{\text{-1.43}}$\\
\textit{+ MLAfication} &  & -68.75\%   & 55.03$_{\text{-0.39}}$& 38.14 & 58.72 & 75.63 & 60.52 & 42.60 & 54.54 &  \textbf{15.59}$_{\text{-1.17}}$\\
\hline  
\textit{+ CPT} & \multirow{3}{*}{700M} & -- &  55.71 & 38.80 & 58.72 & 76.06 & 61.98 & 43.60 & 55.09 & 16.09 \\
\textit{+ TransMLA} &  & -81.25\%   & \textbf{54.78}$_{\text{-0.93}}$ & 37.79 & 57.53 & 75.52 & 59.88 & 42.80 & 55.17 & 14.82$_{\text{-1.27}}$\\
\textit{+ MLAfication} &  & -81.25\%   & 54.42$_{\text{-1.29}}$ & 37.10 & 56.28 & 75.46 & 59.68 & 44.00 & 53.99 & \textbf{15.39}$_{\text{-0.70}}$\\
\hline   
\textit{+ CPT} & \multirow{3}{*}{1B} & -- & 55.84 & 39.15 & 59.38 & 75.63 & 62.26 & 43.00 & 55.64  & 16.49\\
\textit{+ TransMLA} &  & -87.50\%  & \textbf{54.01}$_{\text{-1.83}}$ & 37.24 & 56.32 & 74.81 & 60.08 & 42.40 & 53.20 & \textbf{14.85}$_{\text{-1.64}}$\\
\textit{+ MLAfication} &  & -87.50\%   & 53.95$_{\text{-1.89}}$ & 37.67 & 55.58 & 75.46 & 59.12 & 42.40 & 53.43 & 14.48$_{\text{-2.01}}$\\

\arrayrulecolor{black}
\midrule
\rowcolor{blue!10} \multicolumn{11}{c}{\textit{\textbf{GQA foundation models}}} \\
\rowcolor{gray!10}\textbf{Qwen2.5-7B} & \textcolor{gray}{18T} & \textcolor{gray}{--} & \textcolor{gray}{63.36} & \textcolor{gray}{47.21} & \textcolor{gray}{64.40} & \textcolor{gray}{79.76} & \textcolor{gray}{77.48} & \textcolor{gray}{45.40} & \textcolor{gray}{65.90} & \textcolor{gray}{29.12}\\
\arrayrulecolor{gray!30}
\textit{+ CPT} & \multirow{3}{*}{50M} & -- & 64.13 & 48.24 & 66.27 & 80.09 & 76.55 & 46.20 & 67.40 & 29.14\\
\textit{+ TransMLA} &  & -43.75\%  & 63.33$_{\text{-0.80}}$ & 46.60 & 68.84 & 79.49 & 74.66 & 46.40 & 64.01 & 27.69$_{\text{-1.45}}$\\
\textit{+ MLAfication} &  & -43.75\% & \textbf{63.63}$_{\text{-0.50}}$ & 47.22 & 68.00 & 79.33 & 76.27 & 45.80 & 65.19 & \textbf{29.28}$_{\text{+0.14}}$\\
\hline  
\textit{+ CPT} & \multirow{3}{*}{600M} & -- & 64.42 & 47.94 & 69.98 & 80.30 & 76.02 & 46.00 & 66.30 & 28.90\\
\textit{+ TransMLA} &  & -68.75\%  & 61.88$_{\text{-2.54}}$ & 45.31 & 66.72 & 79.60 & 72.33 & 43.80 & 63.54 & 27.49$_{\text{-1.41}}$\\
\textit{+ MLAfication} &  & -68.75\% & \textbf{62.89}$_{\text{-1.53}}$ & 46.39 & 68.17 & 78.94 & 74.13 & 44.80 & 64.88 & \textbf{28.93}$_{\text{+0.03}}$ \\
\hline   
\textit{+ CPT} & \multirow{3}{*}{1.2B} & -- & 64.11 & 47.96 & 69.55 & 79.87 & 75.80 & 45.20 & 66.30 & 29.36 \\
\textit{+ TransMLA} &  & -81.25\%  & 59.63$_{\text{-4.48}}$ & 43.36 & 63.31 & 78.78 & 70.25 & 42.00 & 60.06 & 26.02$_{\text{-3.34}}$\\ 
\textit{+ MLAfication} &  & -81.25\% & \textbf{60.36}$_{\text{-3.75}}$ & 44.11 & 62.98 & 78.45 & 71.68 & 44.40 & 60.54 & \textbf{27.27}$_{\text{-2.09}}$\\

\arrayrulecolor{black}
\midrule
\rowcolor{blue!10} \multicolumn{11}{c}{\textit{\textbf{GQA+QKNorm}}} \\
\rowcolor{gray!10}\textbf{Qwen3-8B} & \textcolor{gray}{36T} & \textcolor{gray}{--}  & \textcolor{gray}{64.15} & \textcolor{gray}{50.14} & \textcolor{gray}{68.71} & \textcolor{gray}{79.22} & \textcolor{gray}{77.01} & \textcolor{gray}{43.60} & \textcolor{gray}{66.22} &  \textcolor{gray}{30.96}\\
\arrayrulecolor{gray!30}
\textit{+ CPT} & \multirow{2}{*}{300M} & -- & 65.05 & 50.93 & 71.47 & 79.92 & 75.88 & 45.40 & 66.69 & 31.45 \\
\textit{+ MLAfication} &  & -43.75\%   & \textbf{64.27}$_{\text{-0.78}}$ & 50.43 & 72.42 & 79.87 & 75.35 & 43.40 & 64.17 & \textbf{30.83}$_{\text{-0.62}}$ \\
\hline  
\textit{+ CPT} & \multirow{2}{*}{600M} & -- & 64.97 & 50.58 & 71.30 & 80.30 & 75.72 & 45.00 & 66.93 & 31.50 \\
\textit{+ MLAfication} &  & -68.75\%   & \textbf{62.98}$_{\text{-1.99}}$ & 48.68 & 70.15 & 79.54 & 73.75 & 43.40 & 62.35  & \textbf{31.06}$_{\text{-0.44}}$\\
\hline   
\textit{+ CPT} & \multirow{2}{*}{1.2B} & -- & 65.09 & 50.88 & 71.96 & 80.20 & 75.82 & 44.80 & 66.85 & 31.44 \\
\textit{+ MLAfication} &  & -81.25\%   & \textbf{61.74}$_{\text{-3.35}}$ & 46.75 & 67.83 & 78.24 & 72.29 & 44.40 & 60.93 &  \textbf{29.32}$_{\text{-2.12}}$\\

\arrayrulecolor{black}
\midrule
\rowcolor{blue!10} \multicolumn{11}{c}{\textit{\textbf{GQA+QKNorm+MoE}}} \\
\rowcolor{gray!10}\textbf{Qwen3-30B-A3B} & \textcolor{gray}{36T} & \textcolor{gray}{--}  & \textcolor{gray}{65.97}& \textcolor{gray}{53.48} & \textcolor{gray}{68.74} & \textcolor{gray}{80.96} & \textcolor{gray}{79.63} & \textcolor{gray}{45.60} & \textcolor{gray}{67.40} &  \textcolor{gray}{32.65}\\
\arrayrulecolor{gray!30}
\textit{+ CPT} & \multirow{2}{*}{300M} & -- & 66.75 & 54.11 & 71.13 & 81.61 & 78.01 & 46.80 & 68.82 & 30.68\\
\textit{+ MLAfication} &  & -43.75\%   &\textbf{65.35}$_{\text{-1.40}}$ &51.84  & 70.26 & 81.28 & 77.57 & 45.40 & 65.75 &  \textbf{31.36}$_{\text{+0.68}}$\\
\hline  
\textit{+ CPT} & \multirow{2}{*}{1.2B} & -- & 66.25 & 52.85 & 70.20 &  81.99& 78.51 & 46.60 &67.32  & 29.67\\
\textit{+ MLAfication} &  & -68.75\%   &\textbf{64.65}$_{\text{-1.60}}$ & 49.91 & 70.77 & 81.18 & 76.82 & 44.00 & 65.19 & \textbf{29.54}$_{\text{-0.13}}$ \\
\hline   
\textit{+ CPT} & \multirow{2}{*}{2B} & -- & 66.88 & 54.27 & 70.36 & 81.83 & 78.37 & 47.40 & 69.06 & 30.05\\
\textit{+ MLAfication} &  & -81.25\%   & \textbf{62.83}$_{\text{-4.05}}$& 47.59 & 67.03 & 81.01 &  74.32&  43.80& 63.22 & \textbf{28.01}$_{\text{-2.04}}$ \\

\arrayrulecolor{black}
\bottomrule
\end{tabular}
\caption{
Main results across MHA, GQA, QKNorm-based GQA, and QKNorm-based MoE models.
Avg@ and Avg@LB denote the averages over short-context benchmarks and LongBench, respectively.
\textbf{Tokens} denotes recovery-training tokens, and \textbf{KV Mem.}\ denotes the reduction from the original KV cache.
Subscripts report absolute score differences from the CPT baseline.
}
\label{tab:main}
\end{table*}

%% file: tab/qk_norm.tex
\begin{table}[!t]

\centering
\small
\begin{tabular}{@{}lccc@{}}
\toprule
\textbf{Model} & \textbf{KV Mem.} & \textbf{Avg@CS} & \textbf{Avg@LB} \\
\midrule
\rowcolor{gray!10} \textbf{Qwen3-8B} & -- &64.15 &30.96 \\
\arrayrulecolor{gray!30}
\textit{w. KNorm removal} & \multirow{2}{*}{-43.75\%} & 63.40& 30.60\\
\textit{w. Ours}          &                        & \textbf{64.27} & \textbf{30.83}\\
\hline
\textit{w. KNorm removal} & \multirow{2}{*}{-68.75\%} &62.16 & 30.09\\
\textit{w. Ours}          &                         &\textbf{62.98} &\textbf{31.06} \\
\hline
\textit{w. KNorm removal} & \multirow{2}{*}{-81.25\%} &60.84 & 28.31\\
\textit{w. Ours}          &                         &\textbf{61.74} &\textbf{29.32} \\
\arrayrulecolor{black}
\midrule
\arrayrulecolor{gray!30}
\rowcolor{gray!10} \textbf{Qwen3-30B-A3B} & -- &65.97 &32.65 \\
\textit{w. KNorm removal} & \multirow{2}{*}{-43.75\%} &58.46 & 21.95\\
\textit{w. Ours}          &                         &\textbf{65.35} &\textbf{31.36} \\
\hline
\textit{w. KNorm removal} & \multirow{2}{*}{-68.75\%} &61.23 &26.03 \\
\textit{w. Ours}          &                         &\textbf{64.65} &\textbf{29.54} \\
\hline
\textit{w. KNorm removal} & \multirow{2}{*}{-81.25\%} &59.47 &23.71 \\
\textit{w. Ours}          &                         &\textbf{62.83} &\textbf{28.01} \\
\arrayrulecolor{black}
\bottomrule
\end{tabular}
\caption{
Effect of QKNorm adaptation.
Compared with directly removing KNorm, our method consistently improves downstream performance
under different KV-cache reduction ratios.
}
\label{tab:qknorm_ablation}
\end{table}

%% file: tab/quant_30B.tex
\begin{table*}[!t]

\centering
\definecolor{cellblue}{HTML}{e7f5ff}
\definecolor{textblue}{HTML}{175bb5}
\setlength{\tabcolsep}{5pt}
\renewcommand{\arraystretch}{1.10}

\begin{tabular}{lrl@{\hspace{8pt}}rl@{\hspace{8pt}}rl@{\hspace{8pt}}rl}
\toprule
\multirow{2}{*}{\textbf{Precision}}
& \multicolumn{2}{c}{$\bm{r=/}$}
& \multicolumn{2}{c}{$\bm{r=512}$}
& \multicolumn{2}{c}{$\bm{r=256}$}
& \multicolumn{2}{c}{$\bm{r=128}$} \\
\cmidrule(lr){2-3}
\cmidrule(lr){4-5}
\cmidrule(lr){6-7}
\cmidrule(lr){8-9}
& \textbf{KV Mem.} & \textbf{Avg@LB}
& \textbf{KV Mem.} & \textbf{Avg@LB}
& \textbf{KV Mem.} & \textbf{Avg@LB}
& \textbf{KV Mem.} & \textbf{Avg@LB} \\
\midrule

BF16
& 100.00\% & {32.65}
& -43.75\% & {31.36}$_{-1.29}$
& -68.75\% & {29.54}$_{-3.11}$
& -81.25\% & {28.01}$_{-4.64}$ \\

FP8$_{\mathrm{Quanto}}$
& -50.00\% & 31.97$_{-0.68}$
& -71.88\% & 29.31$_{-3.34}$
& -84.38\% & 28.98$_{-3.67}$
& -90.63\% & 26.90$_{-5.75}$ \\

\arrayrulecolor{gray!50}
\hline
\arrayrulecolor{black}

Int4$_{\mathrm{HQQ}}$
& \multirow{2}{*}{-71.88\%}
& 31.76$_{-0.89}$
& \multirow{2}{*}{\textcolor{textblue}{-84.18\%}}
& 29.19$_{-3.46}$
& \multirow{2}{*}{\textcolor{textblue}{-91.21\%}}
& 28.63$_{-4.02}$
& \multirow{2}{*}{\textcolor{textblue}{-94.73\%}}
& 27.00$_{-5.65}$ \\

Int4$_{\mathrm{Quanto}}$
& & 31.47$_{-1.18}$
& & \cellcolor{cellblue}29.28$_{-3.37}$
& & \cellcolor{cellblue}28.89$_{-3.76}$
& & \cellcolor{cellblue}27.21$_{-5.44}$ \\

\arrayrulecolor{gray!50}
\hline
\arrayrulecolor{black}

Int2$_{\mathrm{HQQ}}$
& \multirow{2}{*}{\textcolor{textblue}{-84.38\%}}
& \cellcolor{cellblue}28.97$_{-3.68}$
& \multirow{2}{*}{-91.21\%}
& 25.11$_{-7.54}$
& \multirow{2}{*}{-95.12\%}
& 24.89$_{-7.76}$
& \multirow{2}{*}{-97.07\%}
& 21.70$_{-10.95}$ \\

Int2$_{\mathrm{Quanto}}$
& & 10.92$_{-21.73}$
& & 23.33$_{-9.32}$
& & 23.22$_{-9.43}$
& & 18.80$_{-13.85}$ \\

\bottomrule
\end{tabular}
\caption{LongBench results of Qwen3-30B-A3B-base under different latent dimensions and KV-cache precisions. KV-memory reductions are measured against the original BF16 cache; with a group size of 64, quantization results in effective precisions of 4.5 bits for Int4 and 2.5 bits for Int2. Subscripts denote absolute score differences from the original BF16 baseline (32.65).}
\label{tab:30b-a3b-quant-kv}
\end{table*}

%% file: tab/ablation_at_distill.tex
\begin{table}[!t]

\centering
\small
\setlength{\tabcolsep}{3pt}
\begin{tabular}{clccc}
\toprule
$\mathcal{L}$ & \textbf{Fine-tuning} & \textbf{KV Mem.} & \textbf{Avg@CS} & \textbf{Avg@LB} \\
\midrule

\rowcolor{gray!10} \multicolumn{5}{c}{\textit{\textbf{SmolLM-1.7B-MLA}}} \\
\arrayrulecolor{gray!30}
\multirow{2}{*}{NTP} & \textit{Full} & \multirow{4}{*}{-68.75\%} & 54.69 & 14.53 \\
                         & \textit{QKV-only}  &                           & 54.51 & 13.72 \\
\multirow{2}{*}{Distill} & \textit{Full} &                           & 54.38 & \textbf{16.54} \\
                         & \textit{QKV-only}  &                           & \textbf{55.03} & 15.59 \\

\hline

\multirow{2}{*}{NTP}     & \textit{Full} & \multirow{4}{*}{-81.25\%} & 53.43 & 13.55 \\
                         & \textit{QKV-only}  &                           & 53.31 & 13.67 \\
\multirow{2}{*}{Distill} & \textit{Full} &                           & 53.26 & \textbf{15.72} \\
                         & \textit{QKV-only}  &                           & \textbf{54.42} & 15.39 \\
\hline

\multirow{2}{*}{NTP}     & \textit{Full} & \multirow{4}{*}{-87.50\%} & 52.88 & 12.92 \\
                         & \textit{QKV-only}  &                           & 52.35 & 12.96 \\
\multirow{2}{*}{Distill} & \textit{Full} &                           & 53.41 & \textbf{15.87} \\
                         & \textit{QKV-only}  &                           & \textbf{53.95} & 14.48 \\
\arrayrulecolor{black}

\midrule

\rowcolor{gray!10} \multicolumn{5}{c}{\textit{\textbf{Qwen3-8B-MLA}}} \\
\arrayrulecolor{gray!30}

\multirow{2}{*}{NTP}     & \textit{Full} & \multirow{4}{*}{-43.75\%} & 63.57 &  28.07\\
                         & \textit{QKV-only}  &                        & 63.83 &  28.73\\
\multirow{2}{*}{Distill} & \textit{Full} &                        & 63.24 &  \textbf{31.05}\\
                         & \textit{QKV-only}  &                        & \textbf{64.27} & 30.83 \\
\hline

\multirow{2}{*}{NTP}     & \textit{Full} & \multirow{4}{*}{-68.75\%} & 61.84 & 26.73 \\
                         & \textit{QKV-only}  &                        & 62.38 & 26.58 \\
\multirow{2}{*}{Distill} & \textit{Full} &                        & 62.28 & \textbf{31.79}\\
                         & \textit{QKV-only}  &                        & \textbf{62.98} & 31.06 \\
\hline

\multirow{2}{*}{NTP}     & \textit{Full} & \multirow{4}{*}{-81.25\%} & 61.00 &  26.15\\
                         & \textit{QKV-only}  &                        & 60.97 & 26.06 \\
\multirow{2}{*}{Distill} & \textit{Full} &                        & 61.30 & \textbf{30.29} \\
                         & \textit{QKV-only}  &                        & \textbf{61.74} & 29.32 \\
\arrayrulecolor{black}

\bottomrule
\end{tabular}
\caption{Ablation of distillation and QKV-only fine-tuning.}
\label{tab:ablation_distill}
\end{table}

%% file: tab/step0_attn_output_mse.tex
\begin{table}[!htbp]

\centering
\small
\begin{tabular}{llcc}
\toprule
\textbf{Stage} & $\Delta\mathcal{M}$ & \textbf{KV Mem.} & \textbf{Attn. MSE} $\downarrow$ \\
\midrule
\rowcolor{gray!10}\multicolumn{4}{c}{\textit{\textbf{Low-rank KV Approximation}}} \\
\arrayrulecolor{gray!30}
MHA2MLA & \textit{w. SVD} & \multirow{2}{*}{-68.75\%} & 482.43 \\
MLAfication & \textit{w. SVDLLM V2} & & \textbf{86.58} \\
\hline
MHA2MLA & \textit{w. SVD} & \multirow{2}{*}{-81.25\%} & 866.92 \\
MLAfication & \textit{w. SVDLLM V2} & & \textbf{244.69} \\
\hline
MHA2MLA & \textit{w. SVD} & \multirow{2}{*}{-87.50\%} & 1189.58 \\
MLAfication & \textit{w. SVDLLM V2} & & \textbf{442.26} \\
\arrayrulecolor{black}
\bottomrule
\end{tabular}
\caption{
Attention-editing ablation on Qwen2.5-7B, measured by the average layer-wise MSE between original and edited attention outputs before recovery.}
\label{tab:step0_attn_output_mse}
\end{table}

%% file: tab/zero_shot_commonsense.tex
\begin{table}[!t]

\centering
\small
\begin{tabular}{@{}lccc@{}}
\toprule
\textbf{Model} & \textbf{KV Mem.} & \textbf{Avg@CS} & \textbf{Avg@LB} \\
\midrule
\arrayrulecolor{gray!30}
\rowcolor{gray!10} \textbf{LLaMA-2-7B} & -- & \text{59.85} & \text{27.40} \\
\textit{+ TransMLA}    & \multirow{2}{*}{-68.75\%} & \textcolor{gray}{58.20} & \textcolor{red}{\hphantom{0}1.34} \\
\textit{+ MLAfication} &                           & \textcolor{gray}{52.68} & \textcolor{red}{\hphantom{0}1.06} \\
\hline
\textit{+ TransMLA}    & \multirow{2}{*}{-87.50\%} & \textcolor{gray}{51.19} & \textcolor{red}{\hphantom{0}0.95} \\
\textit{+ MLAfication} &                           & \textcolor{gray}{42.89} & \textcolor{red}{\hphantom{0}1.19} \\
\hline
\textit{+ TransMLA}    & \multirow{2}{*}{-92.97\%} & \textcolor{gray}{43.26} & \textcolor{red}{\hphantom{0}0.74} \\
\textit{+ MLAfication} &                           & \textcolor{gray}{34.43} & \textcolor{red}{\hphantom{0}0.84} \\
\arrayrulecolor{black}
\midrule
\arrayrulecolor{gray!30}
\rowcolor{gray!10} \textbf{Qwen2.5-7B} & -- & 63.36 & 29.12 \\
\textit{+ TransMLA}    & \multirow{2}{*}{-43.75\%} & \textcolor{gray}{61.38} & \textcolor{red}{\hphantom{0}4.08} \\
\textit{+ MLAfication} &                           & \textcolor{gray}{60.44} & \textcolor{red}{\hphantom{0}7.48} \\
\hline
\textit{+ TransMLA}    & \multirow{2}{*}{-68.75\%} & \textcolor{gray}{52.34} & \textcolor{red}{\hphantom{0}1.76} \\
\textit{+ MLAfication} &                           & \textcolor{gray}{50.51} & \textcolor{red}{\hphantom{0}4.94} \\
\hline
\textit{+ TransMLA}    & \multirow{2}{*}{-81.25\%} & \textcolor{gray}{34.97} & \textcolor{red}{\hphantom{0}0.82} \\
\textit{+ MLAfication} &                           & \textcolor{gray}{34.33} & \textcolor{red}{\hphantom{0}3.33} \\
\arrayrulecolor{black}
\bottomrule
\end{tabular}
\caption{Zero-shot TransMLA vs. MLAfication.}
\label{tab:zero-shot}
\end{table}

%% file: tab/moe_load_balance.tex
\begin{table}[!t]

\centering
\small
\begin{tabular}{lccc}
\toprule
\textbf{Method} & \textbf{KV Mem.} & \textbf{Max Vio.} & \textbf{Mean Vio.} \\
\midrule
Baseline & -- & 5.48 & 2.59 \\
\midrule
\multirow{3}{*}{Ours}
& -43.75\% & 6.04 & 2.70 \\
& -68.75\% & 5.74 & 2.69 \\
& -81.25\% & 5.92 & 2.73 \\
\midrule
\multirow{3}{*}{KNorm removal}
& -43.75\% & 10.53 & 4.21 \\
& -68.75\% & 12.48 & 3.28 \\
& -81.25\% & 11.77 & 3.24 \\
\bottomrule
\end{tabular}
\caption{
MoE load-balancing violations on Qwen3-30B-A3B. Max Vio. and Mean Vio. denote the maximum and mean of the layer-wise maxima, respectively.}
\label{tab:moe_load_balance}
\end{table}

%% file: sec/5_related_work.tex
\section{Related Work}
\label{sec:related-work}

\paragraph{Post-training migration to MLA.}
MLA caches a low-dimensional KV tensor for efficient
decoding~\citep{corr/abs-2405-04434}.
MHA2MLA converts pretrained MHA/GQA using Partial-RoPE and joint
low-rank approximation~\citep{ji-etal-2025-towards}, while TransMLA improves
positional conversion and KV factorization~\citep{conf/nips/MengTYSZ25}.
MHA2MLA-VLM extends contribution-aware MLA conversion to vision-language
models~\citep{conf/aaai/FanSJSG26}.
MLAfication builds on contribution-aware Partial-RoPE and SVD-LLM V2
initialization~\citep{wang-etal-2025-svd-llm}, focusing on QKNorm compatibility
and low-cost recovery for dense or MoE models.

\paragraph{Low-rank KV-cache compression.}
Palu compresses projected keys and values through low-rank
factorization~\citep{chang2024palu}; TALE and SALS further adapt rank or
sparsity to token importance~\citep{lee-etal-2025-tale,mu2025sals}.
MLAfication instead produces an MLA parameterization directly consumable by
optimized latent-attention kernels.

\paragraph{Orthogonal cache compression.}
KV-cache quantization reduces numerical precision~\citep{liu2024kivituningfreeasymmetric2bit,
hooper2025kvquant10millioncontext,li2025kvtuner}, while token selection and
cross-layer sharing reduce the sequence or layer dimension~\citep{nips/LiHYVLYCLC24,
nips/LiuLPHHZ24,corr/abs-2406-09297}.
These methods complement architectural conversion, motivating our evaluation
of MLAfication with low-precision KV caches.

%% file: sec/6_conclusion.tex
\section{Conclusion}
We demonstrate that pretrained MHA/GQA checkpoints can be migrated to MLA with controlled quality loss across a broad range of KV-cache budgets.
MLAfication achieves this consistently across standard attention, QKNorm, and Mixture-of-Experts architectures through architecture-aware conversion and targeted recovery.
The resulting cache reductions translate into substantial serving gains and remain compatible with low-precision KV-cache formats.
Together, these findings make post-training MLA conversion a practical way to improve the inference efficiency of existing LLMs without retraining them from scratch.

\section*{Acknowledgments}
This work was partially funded by Grant No. JCKY2024206A015, the National Natural Science Foundation of China (Grant No. 62506079), and the Fudan Kunpeng \& Ascend Center of Cultivation. The computations for this research were performed on the CFFF platform at Fudan University.

%% file: sec/8_appendix.tex
\section{Training Configuration}
\label{sec:appendix_training}

We follow MHA2MLA and TransMLA in reporting recovery hyperparameters by model and KV-cache reduction ratio.
All runs use the MHA2MLA data mixture, i.e., FineWeb-Edu-Dedup, Cosmopedia-v2, Python-Edu, Open-Web-Math, and StackOverflow.
The recovery setting combines logit distillation from the original model with QKV-only fine-tuning.
For contribution-aware partial rope scoring, we use 128 calibration sequences of 4K tokens each.
QKV-only fine-tuning updates QKV projections and attention normalization parameters.
Unless otherwise specified, we use a global batch size (GBS) of 32 on H200 GPUs and warm up over 3\% of the training data.
Table~\ref{tab:appendix_setup} summarizes the model-specific recovery hyperparameters.

\begin{table}[h]
\caption{Recovery hyperparameters for MLAfication. The trainable parameters are restricted to QKV projections and attention normalization parameters. Approximate trainable ratios are 17.6\% for SmolLM-1.7B, 6.1\% for Qwen2.5-7B, 11.1\% for Qwen3-8B, and 1.6\% for Qwen3-30B-A3B.}
\label{tab:appendix_setup}
\centering
\small
\setlength{\tabcolsep}{4pt}
\renewcommand{\arraystretch}{1.08}
\begin{tabular}{lccccc}
\toprule
\textbf{Model} & \textbf{KV Mem.} & $\bm{d_{kv}}$ & \textbf{Tokens} & \textbf{LR} & \textbf{L} \\
\midrule
\rowcolor{gray!10}
SmolLM-1.7B & -68.75\% & 1024 & 300M & $1e{-4}$ & 2K \\
SmolLM-1.7B & -81.25\% & 640 & 700M & $5e{-4}$ & 2K \\
SmolLM-1.7B & -87.50\% & 384  & 1B   & $5e{-4}$ & 2K \\
\midrule
\rowcolor{gray!10}
Qwen2.5-7B & -43.75\% & 512 & 50M  & $1e{-4}$ & 4K \\
Qwen2.5-7B & -68.75\% & 256 & 600M & $1e{-4}$ & 4K \\
Qwen2.5-7B & -81.25\% & 128 & 1.2B & $1e{-4}$ & 4K \\
\midrule
\rowcolor{gray!10}
Qwen3-8B & -43.75\% & 1024 & 300M & $1e{-4}$ & 4K \\
Qwen3-8B & -68.75\% & 512  & 600M & $1e{-4}$ & 4K \\
Qwen3-8B & -81.25\% & 256  & 1.2B & $1e{-4}$ & 4K \\
\midrule
\rowcolor{gray!10}
Qwen3-30B-A3B & -43.75\% & 512 & 300M & $1e{-4}$ & 4K \\
Qwen3-30B-A3B & -68.75\% & 256 & 1.2B & $1e{-4}$ & 4K \\
Qwen3-30B-A3B & -81.25\% & 128 & 2B   & $1e{-4}$ & 4K \\
\bottomrule
\end{tabular}
\end{table}

\section{Evaluation Configuration}
\label{sec:appendix_evaluation}

\paragraph{Hardware and software.}
We run all serving experiments on a single NVIDIA H200 GPU with
143{,}771 MiB of memory. Each experiment uses one GPU with tensor
parallel size 1, thereby measuring single-GPU serving efficiency without
cross-device communication. The system uses NVIDIA driver 570.133.20,
CUDA compiler version 12.9, and CUDA runtime 12.8. The software stack
consists of PyTorch 2.11.0+cu128 and vLLM 0.19.2.

\paragraph{Models and numerical formats.}
We benchmark Qwen2.5-7B and Qwen3-30B-A3B to cover both dense and MoE
architectures. Model weights remain in BF16 for all experiments.
For MLAfication, we evaluate latent dimensions
$d_c\in\{128,256,512\}$ with the rope dimension fixed to $d_r=64$.
The KV cache is stored in either BF16 or FP8 E4M3; therefore, the FP8
setting changes only the KV-cache representation rather than the model
weights.

\paragraph{Serving workloads.}
We use equal prompt and generation lengths of $4096+4096$,
$8192+8192$, and $16384+16384$ tokens. These workloads progressively
increase the pressure on KV-cache capacity and memory bandwidth while
keeping the prefill and decoding lengths balanced. For each sequence
length, we evaluate concurrency levels of 32, 64, 128, and 256.
Prefix caching is disabled so that throughput gains cannot come from
reusing cached prefixes across requests. Decoding uses temperature 0.0.

\paragraph{Measurement and normalization.}
For every combination of model, latent dimension (where applicable),
cache precision, sequence length, and concurrency, we first run one
warmup iteration, which is excluded from measurement, followed by five
timed iterations. We measure output-token
throughput as the total number of generated tokens divided by the
corresponding wall-clock time. To make results comparable across models
and workloads, we normalize each measurement by the throughput of the
original BF16 GQA model evaluated with the same model, sequence length,
and concurrency:
\begin{equation}
\mathrm{NT}
=
\frac{T_{\mathrm{eval}}}{T_{\mathrm{base}}},
\end{equation}
where $T_{\mathrm{eval}}$ is the throughput of the evaluated configuration
and $T_{\mathrm{base}}$ is that of the matched BF16 GQA baseline.

\paragraph{Detailed throughput results.}
Table~\ref{tab:throughput_detailed} reports absolute output-token
throughput for each GQA BF16 baseline and normalized speedups for all
other configurations. Each speedup uses the baseline from the same model,
prompt/output length, and concurrency. All model weights remain in BF16;
BF16 and FP8 in the configuration names denote the KV-cache precision,
with FP8 using the E4M3 format. TransMLA is reported only on Qwen2.5-7B
because it is not applicable to the QKNorm-based Qwen3 models. Under
matched settings, our lighter-weight rope handling yields slightly higher
throughput than TransMLA; entries are omitted where its optimized kernels
do not support the latent size or where QKNorm makes TransMLA
inapplicable.

\begin{table*}[t]
\caption{
Absolute output-token throughput (tokens/s) for the GQA BF16 baselines
and normalized speedups for all other configurations. Here, 4K, 8K, and
16K denote equal prompt and output lengths, and C denotes concurrency.
}
\label{tab:throughput_detailed}
\centering
\scriptsize
\setlength{\tabcolsep}{2.2pt}
\renewcommand{\arraystretch}{1.06}
\resizebox{0.95\textwidth}{!}{%
\begin{tabular}{lrrrrrrrrrrrr}
\toprule
\textbf{Configuration}
& \multicolumn{4}{c}{\textbf{4K}}
& \multicolumn{4}{c}{\textbf{8K}}
& \multicolumn{4}{c}{\textbf{16K}} \\
\cmidrule(lr){2-5}\cmidrule(lr){6-9}\cmidrule(lr){10-13}
& \textbf{C32} & \textbf{C64} & \textbf{C128} & \textbf{C256}
& \textbf{C32} & \textbf{C64} & \textbf{C128} & \textbf{C256}
& \textbf{C32} & \textbf{C64} & \textbf{C128} & \textbf{C256} \\
\midrule
\rowcolor{gray!10}
\multicolumn{13}{c}{\textit{\textbf{Qwen2.5-7B}}} \\
GQA, BF16
& 3942 & 5463 & 6762 & 7272
& 2963 & 3766 & 4301 & 3914
& 1988 & 2303 & 2083 & 2079 \\
GQA, FP8
& $1.09\times$ & $1.14\times$ & $1.16\times$ & $1.20\times$
& $1.17\times$ & $1.20\times$ & $1.22\times$ & $1.41\times$
& $1.23\times$ & $1.26\times$ & $1.50\times$ & $1.39\times$ \\
MLA-512, BF16
& $1.03\times$ & $1.13\times$ & $1.21\times$ & $1.26\times$
& $1.15\times$ & $1.24\times$ & $1.32\times$ & $1.44\times$
& $1.27\times$ & $1.35\times$ & $1.55\times$ & $1.38\times$ \\
TransMLA-512, BF16
& $1.01\times$ & $1.10\times$ & $1.19\times$ & $1.23\times$
& $1.13\times$ & $1.22\times$ & $1.31\times$ & $1.41\times$
& $1.26\times$ & $1.34\times$ & $1.55\times$ & $1.37\times$ \\
MLA-512, FP8
& $1.07\times$ & $1.25\times$ & $1.35\times$ & $1.45\times$
& $1.24\times$ & $1.44\times$ & $1.56\times$ & $1.89\times$
& $1.50\times$ & $1.65\times$ & $2.11\times$ & $2.10\times$ \\
MLA-256, BF16
& $1.16\times$ & $1.33\times$ & $1.49\times$ & $1.61\times$
& $1.37\times$ & $1.59\times$ & $1.78\times$ & $2.18\times$
& $1.66\times$ & $1.88\times$ & $2.47\times$ & $2.30\times$ \\
TransMLA-256, BF16
& $1.08\times$ & $1.19\times$ & $1.26\times$ & $1.33\times$
& $1.22\times$ & $1.33\times$ & $1.42\times$ & $1.68\times$
& $1.39\times$ & $1.47\times$ & $1.85\times$ & $1.72\times$ \\
MLA-256, FP8
& $1.16\times$ & $1.37\times$ & $1.62\times$ & $1.77\times$
& $1.44\times$ & $1.73\times$ & $2.01\times$ & $2.52\times$
& $1.80\times$ & $2.18\times$ & $2.95\times$ & $3.24\times$ \\
MLA-128, BF16
& $1.20\times$ & $1.40\times$ & $1.56\times$ & $1.69\times$
& $1.44\times$ & $1.68\times$ & $1.88\times$ & $2.31\times$
& $1.76\times$ & $2.04\times$ & $2.67\times$ & $2.87\times$ \\
MLA-128, FP8
& $1.22\times$ & $1.47\times$ & $1.76\times$ & $1.98\times$
& $1.50\times$ & $1.88\times$ & $2.30\times$ & $2.93\times$
& $1.96\times$ & $2.50\times$ & $3.56\times$ & $3.98\times$ \\
\midrule
\rowcolor{gray!10}
\multicolumn{13}{c}{\textit{\textbf{Qwen3-30B-A3B MoE}}} \\
GQA, BF16
& 2740 & 3602 & 3351 & 3386
& 1914 & 1809 & 1816 & 1899
& 959 & 922 & 954 & 979 \\
GQA, FP8
& $1.11\times$ & $1.14\times$ & $1.55\times$ & $1.42\times$
& $1.18\times$ & $1.60\times$ & $1.46\times$ & $1.42\times$
& $1.60\times$ & $1.50\times$ & $1.45\times$ & $1.45\times$ \\
MLA-512, BF16
& $1.19\times$ & $1.28\times$ & $1.76\times$ & $1.40\times$
& $1.35\times$ & $1.85\times$ & $1.49\times$ & $1.42\times$
& $1.85\times$ & $1.61\times$ & $1.49\times$ & $1.45\times$ \\
MLA-512, FP8
& $1.30\times$ & $1.49\times$ & $2.01\times$ & $2.35\times$
& $1.53\times$ & $2.25\times$ & $2.59\times$ & $2.17\times$
& $2.22\times$ & $2.86\times$ & $2.43\times$ & $2.41\times$ \\
MLA-256, BF16
& $1.23\times$ & $1.37\times$ & $2.07\times$ & $2.48\times$
& $1.53\times$ & $2.25\times$ & $2.86\times$ & $2.12\times$
& $2.38\times$ & $3.14\times$ & $2.38\times$ & $2.37\times$ \\
MLA-256, FP8
& $1.20\times$ & $1.43\times$ & $2.23\times$ & $2.82\times$
& $1.63\times$ & $2.51\times$ & $3.30\times$ & $3.62\times$
& $2.69\times$ & $3.57\times$ & $4.14\times$ & $3.57\times$ \\
MLA-128, BF16
& $1.48\times$ & $1.75\times$ & $2.45\times$ & $2.86\times$
& $1.80\times$ & $2.66\times$ & $3.34\times$ & $3.23\times$
& $2.80\times$ & $3.56\times$ & $3.66\times$ & $3.00\times$ \\
MLA-128, FP8
& $1.53\times$ & $1.87\times$ & $2.79\times$ & $3.31\times$
& $1.85\times$ & $2.88\times$ & $3.80\times$ & $4.45\times$
& $3.07\times$ & $4.35\times$ & $5.33\times$ & $5.35\times$ \\
\bottomrule
\end{tabular}%
}
\end{table*}

\section{Background: MLA and MHA2MLA}
\label{sec:appendix_mla_background}

\subsection{MHA and GQA}

Given an input token $\bm x_i\in\mathbb R^d$, standard MHA
projects it into a query, key, and value for each attention head
\citep{nips/VaswaniSPUJGKP17}:
\begin{equation}
    \begin{aligned}
        \bm q_i^{(h)} &= \bm W_q^{(h)}\bm x_i,
        &\qquad
        \bm k_i^{(h)} &= \bm W_k^{(h)}\bm x_i,\\
        \bm v_i^{(h)} &= \bm W_v^{(h)}\bm x_i.
    \end{aligned}
\end{equation}
During autoregressive decoding, the keys and values of all previous tokens
must be cached.  MHA therefore stores $2n_h d_h$ elements per token, where
$n_h$ is the number of heads and $d_h$ is the head dimension.  GQA reduces
this cost by sharing each key/value pair among a group of query heads
\citep{emnlp/AinslieLJZLS23}, but still caches full-dimensional keys and
values for every KV head.

\subsection{Multi-Head Latent Attention}

MLA~\citep{corr/abs-2405-04434} replaces the full KV cache with a joint
low-rank representation and separates positional encoding from latent
compression.  Omitting head and group indices for clarity, its
position-agnostic path is
\begin{equation}
    \begin{aligned}
        \bm c_{i,kv} &= \bm W_{dkv}\bm x_i,\\
        \bm k_{i,\mathrm{nope}} &= \bm W_{uk}\bm c_{i,kv},
        &\qquad
        \bm v_i &= \bm W_{uv}\bm c_{i,kv}.
    \end{aligned}
\end{equation}
where the latent dimension $d_c=\dim(\bm c_{i,kv})$ is much smaller than
the total dimension of the original keys and values.  A separate,
low-dimensional path retains RoPE~\citep{journals/ijon/SuALPBL24}:
\begin{equation}
    \begin{aligned}
        \bm q_{i,\mathrm{rope}}
        &= \operatorname{RoPE}(\bm W_{qr}\bm x_i),\\
        \bm k_{i,\mathrm{rope}}
        &= \operatorname{RoPE}(\bm W_{kr}\bm x_i).
    \end{aligned}
\end{equation}
The attention logit combines the two paths:
\begin{equation}
    \begin{aligned}
        s_{ij}={}&
        \left\langle
            \bm q_{i,\mathrm{nope}},
            \bm k_{j,\mathrm{nope}}
        \right\rangle\\
        &+
        \left\langle
            \bm q_{i,\mathrm{rope}},
            \bm k_{j,\mathrm{rope}}
        \right\rangle .
    \end{aligned}
\end{equation}
For the NoPE term,
\begin{equation}
    \begin{aligned}
        &\left\langle
            \bm q_{i,\mathrm{nope}},
            \bm W_{uk}\bm c_{j,kv}
        \right\rangle\\
        &\qquad=
        \left\langle
            \bm W_{uk}^{\top}\bm q_{i,\mathrm{nope}},
            \bm c_{j,kv}
        \right\rangle .
    \end{aligned}
\end{equation}
Thus, the key up-projection can be absorbed into the query projection, and
the value up-projection can likewise be merged with the output projection.
Inference only needs to cache $\bm c_{i,kv}$ and the small RoPE key instead
of the full keys and values.

\begin{table*}[t]
\caption{
Absolute output-token throughput (tokens/s) for the BF16 GQA baseline and
normalized speedups for MLAfication on a single Ascend 910B NPU. Here, 4K,
8K, and 16K denote equal prompt and output lengths, and C denotes concurrency.
}
\label{tab:throughput_ascend}
\centering
\scriptsize
\setlength{\tabcolsep}{2pt}
\renewcommand{\arraystretch}{1.06}
\resizebox{0.95\textwidth}{!}{%
\begin{tabular}{lrrrrrrrrrrrr}
\toprule
\textbf{Configuration}
& \multicolumn{4}{c}{\textbf{4K}}
& \multicolumn{4}{c}{\textbf{8K}}
& \multicolumn{4}{c}{\textbf{16K}} \\
\cmidrule(lr){2-5}\cmidrule(lr){6-9}\cmidrule(lr){10-13}
& \textbf{C16} & \textbf{C32} & \textbf{C64} & \textbf{C128}
& \textbf{C16} & \textbf{C32} & \textbf{C64} & \textbf{C128}
& \textbf{C16} & \textbf{C32} & \textbf{C64} & \textbf{C128} \\
\midrule
\rowcolor{gray!10}
\multicolumn{13}{c}{\textit{\textbf{Qwen2.5-7B on Ascend 910B}}} \\
GQA, BF16
& 695.90 & 1040.95 & 1410.60 & 1417.94
& 546.54 & 736.61 & 765.13 & 788.98
& 382.66 & 396.36 & 412.04 & 426.99 \\
MLA-512, BF16
& $0.834\times$ & $0.924\times$ & $1.014\times$ & $1.362\times$
& $0.933\times$ & $1.069\times$ & $1.407\times$ & $1.326\times$
& $1.084\times$ & $1.458\times$ & $1.373\times$ & $1.410\times$ \\
MLA-128, BF16
& $0.982\times$ & $1.170\times$ & $1.400\times$ & $2.124\times$
& $1.195\times$ & $1.544\times$ & $2.323\times$ & $3.244\times$
& $1.574\times$ & $2.520\times$ & $3.556\times$ & $4.033\times$ \\
\bottomrule
\end{tabular}%
}
\end{table*}

\subsection{MHA2MLA}

MHA2MLA~\citep{ji-etal-2025-towards} converts a pretrained MHA/GQA model
into this MLA form through two steps.  First, \emph{Partial-RoPE} retains
RoPE on a small set of query/key frequency subspaces and removes it from
the remaining dimensions.  The retained subspaces form the position-aware
path, while the resulting NoPE dimensions can be absorbed into the latent
path.  MHA2MLA selects the retained subspaces according to their head-wise
2-norm contribution to the attention logits.

Second, \emph{KV-joint low-rank approximation} jointly factorizes the NoPE
key and value projections:
\begin{equation}
    \begin{bmatrix}
        \bm W_{k,\mathrm{nope}} \\
        \bm W_v
    \end{bmatrix}
    \approx
    \begin{bmatrix}
        \bm W_{uk} \\
        \bm W_{uv}
    \end{bmatrix}
    \bm W_{dkv}.
\end{equation}
A truncated SVD initializes these latent projections, after which recovery
training compensates for the errors introduced by RoPE removal and rank
truncation.  The result preserves the pretrained model's parameters while
making its inference-time KV cache compatible with MLA.

\section{Progressive Compression}
\label{sec:appendix_progressive}

\paragraph{Searching compression ratios.}
Selecting the latent KV dimension is a practical challenge, while independently recovering every candidate is costly.
Table~\ref{tab:stepwise_compression} presents a progressive strategy that starts from a mildly compressed checkpoint, prunes latent dimensions, and continues recovery from the pruned checkpoint.
By reusing earlier recovery, it stays within 0.5 points of independently recovered models at the same target dimension, enabling efficient exploration of the quality--memory frontier.

\input{tab/stepwise_compression}

\section{Ascend Serving Efficiency}
\label{sec:appendix_ascend}

\paragraph{Hardware and execution protocol.}
We additionally evaluate MLAfication on a single Ascend 910B NPU with
CANN 9.1. Each serving process exposes exactly one NPU and uses tensor
parallel size 1. Although the machine contains eight NPUs, they only run
independent benchmark cells in parallel; consequently, the reported
measurements contain no tensor-parallel communication. We use graph-mode
execution, BF16 model weights and KV cache, a block size of 128, and disable
chunked prefill.

\paragraph{Model and workloads.}
The benchmark uses Qwen2.5-7B dummy weights with the exact tensor shapes of
the original model and synthetic token IDs. This setup isolates serving-system
efficiency and does not measure language-model quality. We compare MLAfication
with $d_c\in\{128,512\}$ against the unmodified Qwen2.5-7B GQA architecture
under equal prompt and generation lengths of $4096+4096$, $8192+8192$, and
$16384+16384$ tokens, and concurrency levels of 16, 32, 64, and 128. Each of
the 36 cells contains one timed request batch. Engine initialization and graph
capture occur before the timed interval, whereas the measurement includes
prefill, decode, scheduling, request queueing, and graph replay. Throughput is
normalized by the matched GQA baseline.

\paragraph{Ascend implementation.}
We use graph-updatable PageAttention metadata, fused query/KV projections,
broadcast-mask RoPE, and transpose batch matrix multiplication. CANN accepts
the native 28 heads for $d_c=128$, whereas $d_c=512$ requires padding to 32.
The unsupported $d_c=256$ configuration is omitted.

\paragraph{Results.}
As shown in Table~\ref{tab:throughput_ascend}, MLA-128 outperforms the
baseline in 11 of 12 workloads and achieves a $1.932\times$ geometric-mean
throughput ratio. Its only regression is $1.8\%$ at 4K+4K and C16. MLA-512
wins 9 of 12 workloads with a $1.162\times$ geometric mean; at low load, its
projection cost and mandatory head padding can outweigh its KV-cache savings.
The gains increase as context length and concurrency place more pressure on
KV-cache capacity. At C128, MLA-128 achieves
$2.124\times$--$4.033\times$ throughput across the three sequence lengths,These results show
that the serving gains transfer to Ascend, while also exposing the importance
of latent-width-specific kernel constraints.

%% file: tab/stepwise_compression.tex
\begin{table}[h]

\centering
\small
\begin{tabular}{rcccc}
\toprule
\textbf{Model} & \textbf{Tok.} & \textbf{KV Mem.} & \textbf{CS} & \textbf{LB} \\
\midrule
\arrayrulecolor{gray!30}
\rowcolor{gray!10} \textbf{Qwen2.5-7B} & & --  & 63.36 & 29.12 \\
\singlerightarrow    & 50M & -43.75\%  &  63.63 & 29.28 \\
\hline
\singlerightarrow    & \multirow{2}{*}{600M} & \multirow{2}{*}{-68.75\%}  & 62.89 & 28.93  \\
$_\texttt{+550M}$ \doublerightarrow    &            &               &  62.55 & 28.71 \\
\hline
\singlerightarrow    &\multirow{2}{*}{1.2B} & \multirow{2}{*}{-81.25\%}  & 60.36  & 27.27 \\
$_\texttt{+600M}$ \triplerightarrow &     &                &  59.94 & 27.76 \\
\arrayrulecolor{black}
\bottomrule
\end{tabular}
\caption{Progressive search for KV-cache compression ratios on Qwen2.5-7B.}
\label{tab:stepwise_compression}
\end{table}